\documentclass[runningheads]{llncs}
\usepackage{graphicx}
\usepackage{array}
\usepackage{makecell}
\usepackage{booktabs}
\usepackage{float}
\usepackage{multirow}

\usepackage{subcaption}
\usepackage[T1]{fontenc}
\usepackage[
    colorlinks=true,
    linkcolor=blue,
    citecolor=blue,
    urlcolor=blue
]{hyperref}

\begin{document}
\title{A Multi-Dataset Benchmark of YOLO-Based Weed Detection in Precision Agriculture}
\titlerunning{YOLO-Based Weed Detection Benchmark}

\author{Hristina Zdraveska\inst{1} \and
Vlatko Spasev\inst{1}\orcidID{0009-0008-2650-916X} \and
Ivica Dimitrovski\inst{1}\orcidID{0000-0002-2877-3430}
\and
Ivan Kitanovski\inst{1}\orcidID{0000-0002-0014-4229}
\and
Petre Lameski\inst{1}\orcidID{0000-0002-5336-1796}
}
\authorrunning{H. Zdraveska et al.}
\institute{Faculty of Computer Science and Engineering, University Ss Cyril and Methodius, Rudzer Boshkovikj 16, P.O. 393, Skopje, 1000, North Macedonia}
\maketitle              % typeset the header of the contribution
\begin{abstract}
Weed detection is an important component of precision agriculture, enabling site-specific weed management and reducing unnecessary herbicide use. Although deep learning methods have achieved strong results for crop and weed detection, many studies rely on single-dataset evaluation, making it difficult to assess robustness across different agricultural domains. This paper presents a multi-dataset benchmark of deep object detectors for weed detection in precision agriculture, with a focused evaluation of YOLO26 models. We evaluate nano, small, and medium variants on seven public weed-detection datasets covering different crops, weed species, field conditions, acquisition setups, and annotation protocols. The models are compared in terms of detection accuracy, model complexity, inference latency, FPS, and model size. In addition to in-dataset evaluation, we investigate cross-domain generalization using a unified one-class weed setup and evaluate multi-source training using the combined training subsets from all datasets. The results show that YOLO26 achieves strong in-dataset performance, with YOLO26m obtaining the highest average accuracy and YOLO26s providing the best practical accuracy--efficiency trade-off. However, cross-domain performance decreases substantially, with YOLO26s dropping from an average in-domain mAP$_{50:95}$ of 0.603 to 0.148 in the off-domain setting. Multi-source training improves performance on several datasets, but does not fully eliminate domain shift. Overall, the benchmark highlights the importance of dataset diversity, domain similarity, and target-domain adaptation for robust weed detection in real-world precision agriculture applications.
\end{abstract}

\section{Introduction}

Weeds are among the major biological constraints in agricultural production because they compete with crops for nutrients, water, sunlight, and space~\cite{pai2024deep},~\cite{kamilaris2018deep}. Conventional weed control is commonly based on uniform herbicide application, which can increase production costs and lead to unnecessary chemical use~\cite{milioto2018real},~\cite{lameski2018review}. Precision agriculture aims to reduce this limitation by enabling site-specific treatment, where herbicides or mechanical interventions are applied only where needed. Reliable weed detection is therefore an essential component of site-specific weed management, autonomous weeding, and precision spraying systems. Computer vision has become a central technology for weed detection in field environments~\cite{wu2021review}. Earlier approaches relied on color, vegetation indices, texture, and shape descriptors, while recent methods increasingly use deep learning because it can learn discriminative visual features directly from annotated field imagery. Deep learning has been applied to weed classification, object detection, semantic segmentation, and robotic crop-weed discrimination~\cite{hasan2021survey}. However, robust weed detection remains challenging because field images vary substantially in illumination, soil background, plant growth stage, occlusion, viewpoint, crop type, weed species, and annotation protocol.

Several weed-recognition studies formulate the task as image or patch classification, where the goal is to assign a crop, weed, or weed-species label to an image or cropped plant region~\cite{hasan2021survey}. However, site-specific weed management requires spatial localization, since robotic weeding and precision spraying systems must identify where weeds are located in the field. For this reason, this work focuses on object detection, which predicts both plant categories and bounding-box locations.

Among modern object detectors, YOLO-based models are widely used because they balance detection accuracy and inference speed~\cite{diwan2023object}. This is important for agricultural applications such as robotic weeding and real-time spraying. However, many weed-detection studies are still evaluated on single datasets or under non-unified settings, making it difficult to assess generalization across agricultural domains. Recent public weed-detection datasets provide an opportunity to evaluate object detectors under diverse field conditions, crop types, weed species, growth stages, acquisition setups, and annotation formats~\cite{hasan2021survey}. In this work, we use seven public datasets that represent different agricultural scenarios, including maize fields, cotton production systems, Palmer amaranth detection, mixed crop and weed imagery, and real field environments with varying plant species and backgrounds. This diversity enables evaluation not only under standard in-dataset conditions, but also in cross-domain settings where models trained on one dataset are tested on another.

The main aim of this paper is to provide a multi-dataset benchmark of deep object detectors for weed detection in precision agriculture. We focus on the YOLO26 detector family and evaluate nano, small, and medium variants under a unified experimental protocol. The benchmark includes in-dataset evaluation using the original dataset-specific label spaces, computational efficiency analysis, cross-domain generalization using a unified one-class weed setup, and multi-source training using the combined training subsets from all datasets. This design allows us to analyze not only detection performance under standard in-domain conditions, but also the robustness of weed detectors when applied to unseen agricultural domains.

The main contributions of this work are as follows:

\begin{itemize}
    \item We present a multi-dataset benchmark for weed detection in precision agriculture using seven public datasets with different crops, weed species, field conditions, image-acquisition setups, and annotation protocols.

    \item We evaluate YOLO26 object detectors in nano, small, and medium variants, comparing their detection accuracy, model size, computational complexity, inference latency, and frames per second (FPS).

    \item We analyze in-dataset detection performance using the original label space of each dataset, providing dataset-specific baselines for different weed-detection scenarios.

    \item We investigate cross-domain generalization using a unified one-class weed label space, enabling direct source-target evaluation across datasets with incompatible original taxonomies.

    \item We evaluate multi-source training by combining the training subsets from all datasets and testing the resulting model on each dataset-specific test set.

    \item We provide practical insights into the accuracy-efficiency trade-offs of YOLO26 variants and the effect of domain shift on real-world weed-detection performance.
\end{itemize}

The remainder of the paper is organized as follows. Section~\ref{sec:related_work} reviews related work and provides background on weed detection and YOLO-based object detectors. Section~\ref{sec:materials_methods} describes the datasets included in the benchmark and the evaluated YOLO26 models. Section~\ref{sec:experimental_design} presents the annotation conversion, training protocol, and evaluation metrics. Section~\ref{sec:results} reports and discusses the experimental results, including the in-dataset benchmark, computational efficiency analysis, cross-domain evaluation, and multi-source training. Finally, Section~\ref{sec:conclusion} concludes the paper and outlines directions for future work.

\section{Related Work and Background}
\label{sec:related_work}

Weed detection has been widely studied as part of precision agriculture and site-specific weed management. Traditional approaches commonly relied on vegetation indices, color thresholding, texture descriptors, and handcrafted shape features to separate crops, weeds, and soil background~\cite{wu2021review}. While these methods can be effective under controlled conditions, their performance often decreases in real field environments due to changing illumination, shadows, occlusion, plant overlap, soil variability, and visual similarity between crops and weeds.

Deep learning has become the dominant approach for visual weed detection because it can learn task-specific features directly from annotated data. Convolutional neural networks have been applied to weed classification, crop--weed discrimination, weed-species recognition, plant localization, semantic segmentation, and precision-spraying applications~\cite{hasan2021survey,dang2023yoloweeds,ahmad2021performance}. However, reported results are often difficult to compare directly because existing studies use different datasets, model architectures, annotation formats, training protocols, and evaluation metrics.

Weed recognition can be formulated in different ways depending on the target application. Image classification assigns a single label to an entire image, while patch- or plant-level classification assigns a label to a cropped region or an already isolated plant~\cite{olsen2019deepweeds}. These formulations are useful for evaluating visual separability between crop and weed classes, but they do not directly provide the spatial localization required for site-specific treatment. Semantic and instance segmentation provide more detailed pixel-level or instance-level plant delineation, but require more expensive annotations and may be computationally demanding~\cite{lottes2018fully}, ~\cite{lameski2017weed}. Object detection represents a practical intermediate formulation because it predicts both class labels and bounding boxes for individual plant instances. This makes it suitable for robotic weeding and precision spraying, where the system must determine not only whether weeds are present, but also where they are located.

In the broader object-detection literature, Faster R-CNN represents an influential two-stage detector, while Single-Shot Detector (SSD) is an important single-stage detector~\cite{ren2017faster,liu2016ssd}. YOLO-based detectors are widely used for real-time object detection because of their single-stage design and favorable accuracy--efficiency trade-off~\cite{redmon2016you,allmendinger2025assessing}. Earlier YOLO models demonstrated the feasibility of real-time detection, while more recent versions have introduced improvements in backbone design, feature aggregation, training strategies, and deployment-oriented optimization. These properties make YOLO-based detectors particularly relevant for agricultural applications~\cite{badgujar2024agricultural}, where models may need to operate on embedded devices, agricultural robots, or real-time precision-spraying platforms.

Recent studies have increasingly evaluated YOLO-based detectors for agricultural weed detection. For example, YOLO-based models have been benchmarked for multi-class weed detection in cotton fields~\cite{dang2023yoloweeds}, common weed detection in corn and soybean production systems~\cite{ahmad2021performance}, and real-time weed detection using YOLO- and transformer-based detectors~\cite{allmendinger2025assessing}. Other studies have compared several YOLO generations and Faster R-CNN for detecting multiple weed species~\cite{sharma2024comparative}. These works demonstrate the practical relevance of modern object detectors for weed localization, but most evaluations remain dataset-specific or focus on a limited set of acquisition conditions. As a result, less attention has been given to how detectors behave when transferred across datasets with different crop types, weed species, viewpoints, annotation protocols, and field backgrounds. This gap motivates the multi-dataset and cross-domain evaluation adopted in this work.

Public weed-detection datasets vary significantly in crop type, weed species, acquisition conditions, annotation format, and task definition. Some datasets focus on binary crop and weed detection, while others provide multi-class weed-species annotations or instance-level plant annotations. This diversity is useful for evaluating model robustness, but it also makes direct cross-dataset comparison difficult because the original label spaces and annotation protocols are not always compatible~\cite{zhou2022domain}.

These differences are particularly important when evaluating models intended for deployment in unseen field conditions. A detector trained on one dataset may learn visual patterns that are strongly associated with a specific crop type, soil background, camera viewpoint, growth stage, or annotation style. Therefore, multi-dataset evaluation is necessary to understand not only in-dataset accuracy, but also the robustness of weed detectors under domain shift.

\section{Materials and Methods}
\label{sec:materials_methods}
\subsection{Data Description}
\begin{figure*}
    \centering
    \includegraphics[width=\textwidth]{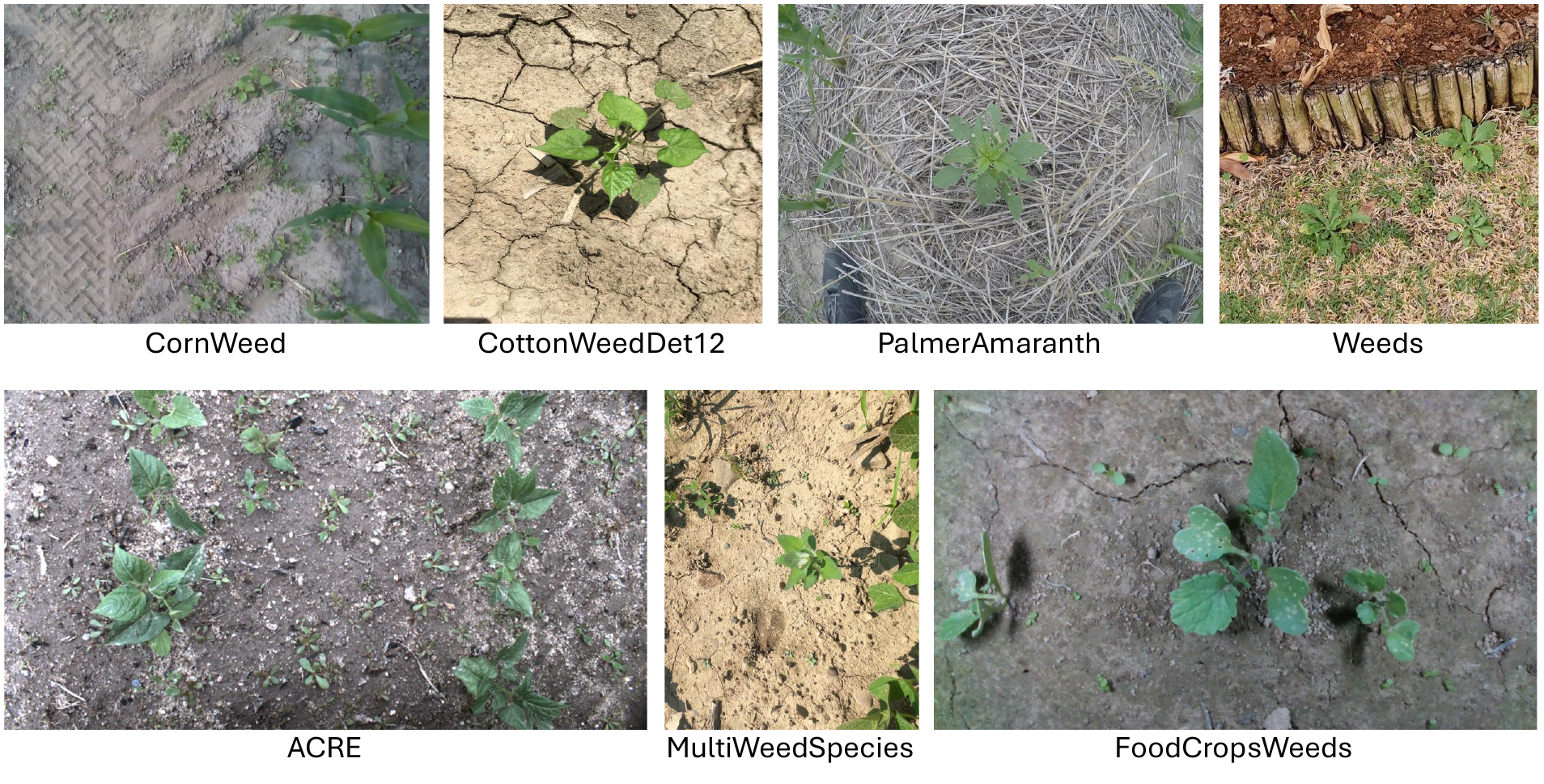}
    \caption{Example images from the seven weed-detection datasets included in the benchmark. The datasets are shown from left to right and top to bottom: CornWeed, CottonWeedDet12, PalmerAmaranth, Weeds, ACRE, MultiWeedSpecies, and FoodCropsWeeds.}
    \label{fig:dataset_examples}
\end{figure*}

This study uses seven public weed-detection datasets selected to cover different crop types, weed species, image-acquisition conditions, and annotation protocols. Table~\ref{tab:weed_detection_datasets} summarizes the main properties of the datasets. Representative examples for each dataset are shown in Figure~\ref{fig:dataset_examples}. 

\newcolumntype{C}[1]{>{\centering\arraybackslash}p{#1}}
\newcommand{\rot}[1]{\rotatebox[origin=c]{90}{\textbf{#1}}}
\begin{table*}
\centering
\caption{Properties of the weed-detection datasets used in the benchmark.}
\label{tab:weed_detection_datasets}
\small
\setlength{\tabcolsep}{6pt}
\renewcommand{\arraystretch}{1.15}
\begin{tabular}{p{2.6cm} C{1.0cm} C{1.0cm} C{0.8cm} C{1.8cm} C{1.0cm} C{0.8cm}}
\toprule
\textbf{Name} &
\rot{\#Images} &
\rot{\#Instances} &
\rot{\#Labels} &
\rot{Image Size} &
\rot{Annotations} &
\rot{Splits} \\
\midrule
CornWeed & 3574 & 281,725 & 2 & \makecell{$640 \times 480$ \\ $1280 \times 720$} & COCO & No \\
CottonWeedDet12 & 5648 & 9388 & 12 & $640 \times 640$ & XML & Yes \\
PalmerAmaranth & 995 & 1405 & 1 & $3000 \times 2250$ & COCO & No \\
FoodCropsWeeds & 1176 & 7853 & 2 & Variable & XML & No \\
ACRE & 1000 & 56530 & 2 & $2046 \times 1080$ & XML & No \\
MultiWeedSpecies & 905 & 1394 & 5 & $4000 \times 3000$ & COCO & No \\
Weeds & 4203 & 11385 & 1 & $416 \times 416$ & COCO & Yes \\
\bottomrule
\end{tabular}
\end{table*}

CornWeed is an object-detection dataset collected in real maize fields for detecting maize plants and weeds~\cite{iqbal_10306233}. It contains 3574 outdoor RGB images annotated with bounding boxes for two classes: maize and weed. The images were acquired from agricultural platforms equipped with an Intel RealSense D435i camera during early maize growth stages. The dataset contains images captured under daylight conditions, including both cloudy and sunny scenes. CottonWeedDet12 is an object-detection dataset collected in cotton fields for detecting multiple weed species in cotton production systems~\cite{dang2023yoloweeds}. It contains 5648 RGB images with bounding-box annotations distributed across 12 weed classes, including waterhemp, morningglory, purslane, spotted spurge, carpetweed, ragweed, eclipta, prickly sida, Palmer amaranth, sicklepod, goosegrass, and cutleaf groundcherry. The images were acquired using handheld smartphones and digital cameras under natural field conditions, covering different weed growth stages, field sites, and lighting conditions.

The Palmer Amaranth dataset is available through the Weed-AI platform and focuses on a single weed species, Palmer amaranth~\cite{weedai2024palmer}. The dataset contains field images annotated with bounding boxes and includes Palmer amaranth plants appearing in different agricultural contexts, including fallow fields, corn, and soybean production systems. This dataset is useful for evaluating single-species weed localization and for studying the effect of crop background variation on detection performance. The Dataset of Annotated Food Crops and Weed Images, denoted as FoodCropsWeeds in this paper, was developed for robotic computer vision control in precision agriculture~\cite{sudars2020dataset}. It contains 1176 RGB images with manually annotated bounding-box objects. Although the dataset includes 8 weed species and 6 food crop species, the object-detection annotations are organized into two primary classes: weed and crop. The images were collected in controlled greenhouse conditions and open-field environments, making the dataset relevant for crop--weed discrimination at early plant growth stages.

The ACRE Crop-Weed dataset was developed for benchmarking weed-detection models in maize and bean fields~\cite{bertoglio2023acre}. The dataset includes two crop types, maize and bean, and four seeded weed species: ryegrass, mustard, matricaria, and lamb's quarter. The images were acquired using a four-wheel skid-steering robot equipped with a downward-facing Basler RGB camera. The original annotations are instance-level plant annotations stored in XML format; for the object-detection experiments, bounding boxes were generated from the instance annotations. The Multiple Weed Species Detection dataset, denoted as MultiWeedSpecies in this paper, is a public Weed-AI dataset containing field imagery annotated for object detection. The dataset contains 905 images and bounding-box annotations for five weed categories: Chenopodium album, Amaranthus palmer, Xanthium strumarium, Taraxacum officinale, and Amaranthus tuberculatus. The images were collected in fallow conditions using a GoPro Hero 8 camera and are provided through Weed-AI in WeedCOCO format~\cite{sharma2024comparative}. Weeds is an object-detection dataset focused on garden weed detection in visually complex backgrounds~\cite{weeds-nxe1w_dataset}. It contains 4,203 images annotated for a single class, weeds. The dataset is designed for object detection scenarios in which weeds can be difficult to distinguish from the surrounding background because of visual similarity with soil, vegetation, and other natural scene elements.

\subsection{YOLO26 Object Detectors}
YOLO is a family of single-stage object detectors designed for real-time object detection. Unlike two-stage detectors, which first generate region proposals and then classify them, YOLO models perform object localization and classification in a single forward pass. This makes them particularly suitable for applications that require fast inference, such as robotic weeding, precision spraying, and embedded agricultural vision systems.

Over time, YOLO architectures have evolved through improvements in backbone design, feature aggregation, training strategies, and deployment efficiency. In this work, we focus on YOLO26, a recent YOLO-based detector designed for real-time and deployment-oriented vision tasks. YOLO26 introduces an end-to-end detection pipeline with native NMS-free inference, reducing the need for a separate non-maximum suppression post-processing step. This is relevant for real-time precision-agriculture applications, where both detection accuracy and inference efficiency are important.

The benchmark evaluates three YOLO26 variants: YOLO26n, YOLO26s, and YOLO26m ~\cite{sapkota2025yolo26}. These variants represent different accuracy--efficiency trade-offs. YOLO26n is the lightweight nano model intended for high-speed inference and resource-constrained deployment. YOLO26s provides a compact configuration with a practical balance between detection accuracy and computational cost. YOLO26m increases model capacity and is included to evaluate whether a larger model improves weed-detection performance. This selection allows the benchmark to analyze how model scale affects detection accuracy, inference speed, model size, and cross-domain generalization.

\section{Experimental Design}
\label{sec:experimental_design}

\subsection{Annotation Conversion and Dataset Preparation}
All datasets were converted to the YOLO object-detection format to ensure consistent training and evaluation. Annotations originally provided in XML, COCO, or WeedCOCO formats were converted to YOLO annotation files. For datasets with segmentation or instance-level annotations, bounding boxes were generated as the minimum enclosing rectangles around annotated plant instances. Invalid, duplicate, or empty annotations were removed during preprocessing.

Two evaluation settings were used. For the in-dataset benchmark, each dataset was evaluated using its original label space, preserving the original task definition, such as crop and weed detection, multi-class weed-species detection, or single-species weed localization. For cross-domain evaluation, all weed categories were mapped to a unified one-class \emph{weed} label space. Crop annotations were removed from the positive label set, but crop plants remained visible in the images; therefore, detections on crop regions were counted as false positives. This setup enables direct source--target evaluation across datasets with different label taxonomies, crop types, weed species, field conditions, and acquisition protocols.

\subsection{Training Protocol}
All YOLO26 models were initialized with pretrained weights and fine-tuned using a fixed input size of $640 \times 640$ pixels. The same training, validation, and test splits were used for all models within each dataset. Official splits were used when available; otherwise, fixed random splits were created. For datasets without predefined splits, images were partitioned using a 70\%--15\%--15\% training, validation, and test ratio. For CottonWeedDet12 and Weeds, the original splits were used.

Each model was trained for a maximum of 100 epochs with early stopping patience set to 20 epochs. The best checkpoint was selected based on validation performance and evaluated on the corresponding test set. All training and inference experiments were conducted on the same workstation equipped with an NVIDIA GeForce GTX 1080 Ti GPU with 11 GB of memory and CUDA 12.2, ensuring consistent latency and FPS comparisons. The evaluated models include three YOLO26 variants: YOLO26n, YOLO26s, and YOLO26m, representing different model scales within the YOLO26 family.

\subsection{Evaluation Metrics}
Detection performance was evaluated using precision, recall, F1-score, mAP$_{50}$, and mAP$_{50:95}$. Precision measures the proportion of predicted detections that are correct, while recall measures the proportion of ground-truth objects that are successfully detected. The F1-score provides a balanced measure of precision and recall. The mAP$_{50}$ metric evaluates detection accuracy at an IoU threshold of 0.50, while mAP$_{50:95}$ follows the COCO-style evaluation protocol and averages mean average precision over IoU thresholds from 0.50 to 0.95~\cite{lin2014coco}.

In addition to detection accuracy, computational efficiency was evaluated using the number of parameters, floating-point operations (FLOPs), model size, inference latency, and frames per second (FPS). These metrics are important for precision-agriculture applications because weed-detection models may need to operate in real time on embedded devices, robotic platforms, or precision-spraying systems. Latency and FPS were measured under the same evaluation setup for all YOLO26 variants to provide a consistent comparison of their practical deployment potential.

\section{Results and Discussion}
\label{sec:results}

\subsection{In-Dataset Benchmark}

The first experiment evaluates the YOLO26 models under in-dataset conditions, where each model is trained and tested on the same dataset using the original dataset-specific label space. This experiment provides the baseline performance for each dataset before evaluating cross-domain generalization and multi-source training. Table~\ref{tab:main_results} reports the detection performance of YOLO26n, YOLO26s, and YOLO26m in terms of precision, recall, F1-score, mAP$_{50}$, and mAP$_{50:95}$.

\begin{table*}
\centering
\caption{In-dataset detection performance of the evaluated YOLO26 models.}
\label{tab:main_results}
\begin{tabular}{l l c c c c c}
\toprule
\textbf{Dataset} & \textbf{Model} & \textbf{Precision} & \textbf{Recall} &
\textbf{F1} & \textbf{mAP$_{50}$} & \textbf{mAP$_{50:95}$} \\
\midrule

\multirow{3}{*}{CornWeed}
& YOLO26n & 0.8662 & 0.8281 & 0.8468 & 0.8878 & 0.5660 \\
& YOLO26s & 0.8703 & 0.8487 & 0.8594 & 0.9027 & 0.5945 \\
& YOLO26m & 0.8791 & 0.8569 & 0.8678 & 0.9063 & \textbf{0.6051} \\
\midrule

\multirow{3}{*}{FoodCropsWeeds}
& YOLO26n & 0.6485 & 0.7573 & 0.6987 & 0.7290 & 0.4831 \\
& YOLO26s & 0.7443 & 0.7717 & 0.7578 & 0.7816 & \textbf{0.5433} \\
& YOLO26m & 0.7304 & 0.8055 & 0.7661 & 0.7888 & 0.5426 \\
\midrule

\multirow{3}{*}{CottonWeedDet12}
& YOLO26n & 0.9157 & 0.8443 & 0.8785 & 0.9046 & 0.8440 \\
& YOLO26s & 0.9365 & 0.8429 & 0.8873 & 0.9116 & 0.8544 \\
& YOLO26m & 0.9289 & 0.8702 & 0.8986 & 0.9254 & \textbf{0.8658} \\
\midrule

\multirow{3}{*}{PalmerAmaranth}
& YOLO26n & 0.8929 & 0.8685 & 0.8806 & 0.9315 & \textbf{0.7332} \\
& YOLO26s & 0.9164 & 0.8826 & 0.8992 & 0.9342 & 0.7227 \\
& YOLO26m & 0.9160 & 0.8709 & 0.8929 & 0.9267 & 0.7311 \\
\midrule

\multirow{3}{*}{ACRE}
& YOLO26n & 0.7466 & 0.6862 & 0.7151 & 0.7298 & 0.4510 \\
& YOLO26s & 0.7509 & 0.7263 & 0.7384 & 0.7637 & 0.4872 \\
& YOLO26m & 0.7565 & 0.7366 & 0.7464 & 0.7689 & \textbf{0.4930} \\
\midrule

\multirow{3}{*}{MultiWeedSpecies}
& YOLO26n & 0.9349 & 0.8271 & 0.8777 & 0.9128 & \textbf{0.6000} \\
& YOLO26s & 0.8988 & 0.8717 & 0.8850 & 0.9109 & 0.5781 \\
& YOLO26m & 0.9146 & 0.8429 & 0.8773 & 0.9092 & 0.5878 \\
\midrule

\multirow{3}{*}{Weeds}
& YOLO26n & 0.9677 & 0.9328 & 0.9499 & 0.9840 & 0.8090 \\
& YOLO26s & 0.9471 & 0.9664 & 0.9566 & 0.9817 & 0.8154 \\
& YOLO26m & 0.9564 & 0.9561 & 0.9563 & 0.9840 & \textbf{0.8175} \\
\bottomrule
\end{tabular}
\end{table*}

The results show that YOLO26 achieves strong in-dataset performance across most datasets, although the absolute performance varies considerably depending on the dataset. The highest mAP$_{50:95}$ values are obtained on CottonWeedDet12 and Weeds, where all three model variants achieve strong detection results. This indicates that the models can learn effective weed representations when the training and test data come from the same visual domain. Strong results are also obtained on PalmerAmaranth, especially considering that it represents a single-species detection task under different crop backgrounds.

The most challenging datasets are ACRE and FoodCropsWeeds. On ACRE, the best mAP$_{50:95}$ is 0.4930, achieved by YOLO26m. This lower performance may be related to the specific acquisition setup, the maize and bean crop context, complex field conditions, and the conversion of instance-level annotations into bounding boxes. FoodCropsWeeds also produces lower scores than CottonWeedDet12, PalmerAmaranth, and Weeds, which may be explained by greater variability in crop and weed appearance, controlled and open-field acquisition conditions, and differences in annotation characteristics.

Increasing model size generally improves in-dataset performance, but the improvement is not uniform across all datasets. YOLO26m achieves the best mAP$_{50:95}$ on CornWeed, CottonWeedDet12, ACRE, and Weeds. However, YOLO26n obtains the best result on PalmerAmaranth and MultiWeedSpecies, while YOLO26s performs best on FoodCropsWeeds. Averaged across the seven datasets, YOLO26n, YOLO26s, and YOLO26m achieve mAP$_{50:95}$ values of approximately 0.6409, 0.6565, and 0.6633, respectively. Thus, YOLO26m provides the highest average accuracy, but the gain over YOLO26s is relatively small. The computational-efficiency results are summarized in Table~\ref{tab:efficiency}. 

\begin{table}
\centering
\caption{Computational efficiency comparison of the evaluated YOLO26 models.
Model size and latency are averaged across the in-dataset test sets.}
\label{tab:efficiency}
\begin{tabular}{lccccc}
\hline
\textbf{Model} &
\textbf{Params (M)} &
\textbf{FLOPs (B)} &
\textbf{Size (MB)} &
\textbf{Latency (ms)} &
\textbf{FPS} \\
\hline
YOLO26n & 2.4  & 5.4  & 5.13  & \textbf{3.21} & \textbf{311.85} \\
YOLO26s & 9.5  & 20.7 & 19.37 & 5.79 & 172.68 \\
YOLO26m & 20.4 & 68.2 & 41.99 & 12.16 & 82.24 \\
\hline
\end{tabular}
\end{table}

As expected, YOLO26n is the most efficient model, with the lowest number of parameters, smallest model size, lowest latency, and highest FPS. YOLO26m achieves the highest average detection accuracy, but it requires substantially more computation: compared with YOLO26s, it has more than twice the number of parameters and model size, more than three times the FLOPs, and approximately twice the inference latency. YOLO26s therefore provides a strong compromise between accuracy and efficiency, achieving accuracy close to YOLO26m while remaining considerably faster and more compact. Based on this trade-off, YOLO26s is used in the subsequent cross-domain and multi-source training experiments. This choice reduces the computational cost of the extended evaluation while preserving competitive detection performance. It also reflects a practical deployment scenario, where the selected model should balance accuracy, inference speed, and model complexity for real-time precision-agriculture applications.

\subsection{Cross-Domain Weed Detection}

\begin{figure*}[t]
    \centering
    \includegraphics[width=\textwidth]{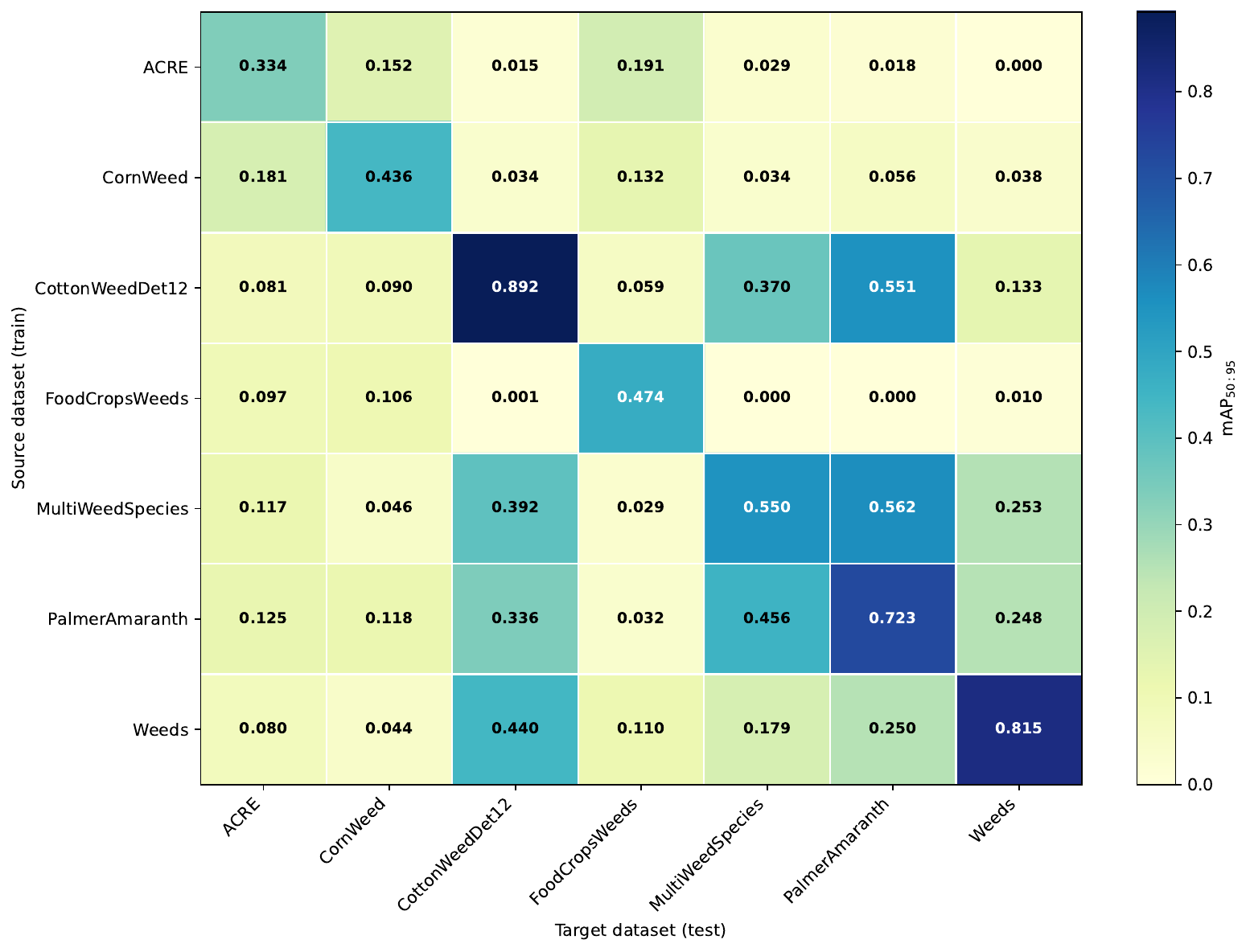}
    \caption{Cross-domain weed-detection performance using the unified one-class weed setup for YOLO26s. Rows indicate the source dataset used for training, while columns indicate the target dataset used for testing. Values report mAP$_{50:95}$.}
    \label{fig:cross_domain_heatmap}
\end{figure*}

Using the harmonized one-class \emph{weed} setup described in Section~\ref{sec:experimental_design}, the cross-domain experiment evaluates whether a model trained on one dataset can generalize to another dataset without fine-tuning. Figure~\ref{fig:cross_domain_heatmap} shows the cross-domain results for YOLO26s. Rows indicate the source dataset used for training, and columns indicate the target dataset used for testing. The diagonal cells represent in-domain evaluation, while the off-diagonal cells represent cross-domain transfer. The results show a clear dominance of the diagonal values, confirming that weed-detection performance is strongly affected by dataset-specific domain characteristics.

The strongest in-domain results are obtained on CottonWeedDet12 and Weeds, with mAP$_{50:95}$ values of 0.892 and 0.815, respectively, while ACRE and FoodCropsWeeds show lower in-domain performance. Cross-domain transfer is substantially weaker. The best transfer is observed from MultiWeedSpecies to PalmerAmaranth, with an mAP$_{50:95}$ of 0.562, followed by CottonWeedDet12 to PalmerAmaranth with 0.551. Strong transfer is also observed from PalmerAmaranth to MultiWeedSpecies and from Weeds to CottonWeedDet12, suggesting that transfer improves when source and target datasets share similar weed appearance, plant morphology, or acquisition conditions.

Overall, the mean in-domain mAP$_{50:95}$ is approximately 0.603, while the mean off-domain mAP$_{50:95}$ is only 0.148, corresponding to a relative reduction of about 75.6\%. This demonstrates that strong in-dataset performance does not necessarily imply robustness on unseen agricultural domains. MultiWeedSpecies, PalmerAmaranth, CottonWeedDet12, and Weeds provide the strongest average off-domain transfer, whereas ACRE and FoodCropsWeeds generalize less effectively. The transfer patterns are also asymmetric, indicating that dataset diversity, weed-species coverage, annotation style, background complexity, and visual similarity all influence cross-domain performance.

\subsection{Multi-Source Training}
The multi-source experiment evaluates whether heterogeneous training data improves robustness across target datasets. A single YOLO26s model was trained on the combined training subsets from all seven datasets and evaluated separately on each dataset-specific test set. This setup complements the cross-domain experiment by testing whether broader source-domain diversity improves generalization. The results are reported in Table~\ref{tab:combined_training_results}. Overall, the multi-source model achieves strong performance on several datasets, with the highest mAP$_{50:95}$ obtained on CottonWeedDet12, followed by Weeds and PalmerAmaranth. The model also achieves competitive results on MultiWeedSpecies and FoodCropsWeeds, indicating that training on a broader set of weed appearances, crop backgrounds, and acquisition conditions can improve performance on several target domains.

However, the results also show that multi-source training does not uniformly improve performance across all datasets. In particular, the performance on ACRE remains relatively low, with an mAP$_{50:95}$ of 0.3241. This suggests that ACRE represents a more challenging target domain, likely due to its specific robot-mounted acquisition setup, maize and bean crop context, lighting variability, and the conversion of instance-level annotations into bounding boxes. Similarly, the result on CottonWeedDet12 is very high but does not substantially exceed the dataset-specific in-domain model, indicating that the additional datasets do not necessarily provide useful information for an already well-represented target domain.

These findings suggest that multi-source training can improve overall robustness, but it does not fully remove the effect of domain shift. The benefit of adding more datasets depends on the similarity and complementarity between the source and target domains. In some cases, additional data improves generalization by increasing visual diversity; in others, it may introduce domain dilution or mild negative transfer. Therefore, multi-source training is useful for building more general weed detectors, but target-domain relevance remains important for achieving optimal performance.

\begin{table*}
\centering
\caption{Performance of the YOLO26s model trained on the combined training sets and evaluated on individual test sets.}
\label{tab:combined_training_results}
\begin{tabular}{l c c c c c}
\toprule
\textbf{Test dataset} & \textbf{Precision} & \textbf{Recall} & \textbf{F1} &
\textbf{mAP$_{50}$} & \textbf{mAP$_{50:95}$} \\
\midrule
Weeds              & 0.9579 & 0.9481 & 0.9530 & 0.9863 & 0.8282 \\
ACRE               & 0.6652 & 0.6322 & 0.6483 & 0.6513 & 0.3241 \\
CornWeed           & 0.8149 & 0.7805 & 0.7973 & 0.8436 & 0.4399 \\
CottonWeedDet12    & 0.9442 & 0.8783 & 0.9100 & 0.9487 & 0.8896 \\
FoodCropsWeeds     & 0.6976 & 0.8292 & 0.7577 & 0.7804 & 0.5231 \\
MultiWeedSpecies   & 0.9056 & 0.8451 & 0.8743 & 0.9069 & 0.5906 \\
PalmerAmaranth     & 0.9463 & 0.8685 & 0.9058 & 0.9403 & 0.7648 \\
\bottomrule
\end{tabular}
\end{table*}

\section{Conclusion}
\label{sec:conclusion}

This paper presented a multi-dataset benchmark of deep object detectors for weed detection in precision agriculture, with a focused experimental evaluation of YOLO26 nano, small, and medium variants. The benchmark considered in-dataset detection performance, computational efficiency, cross-domain generalization, and multi-source training across diverse weed-detection datasets with different crop types, weed species, acquisition conditions, and annotation protocols. The in-dataset experiments showed that YOLO26 models achieve strong detection performance when training and testing data originate from the same dataset. YOLO26m achieved the highest average in-dataset accuracy, with an average mAP$_{50:95}$ of 0.6633 across the evaluated datasets. However, the improvement over YOLO26s was relatively small, while YOLO26m required substantially higher computational cost. YOLO26s therefore provided the most practical accuracy--efficiency trade-off, combining competitive detection accuracy with lower latency, smaller model size, and reduced computational complexity. This makes YOLO26s a suitable candidate for extended generalization experiments and for practical precision-agriculture scenarios where inference speed and resource efficiency are important. The cross-domain evaluation demonstrated that strong in-dataset performance does not necessarily translate to robust performance on unseen agricultural domains. When YOLO26s was trained on one dataset and evaluated on another using a unified one-class \emph{weed} label space, the mean mAP$_{50:95}$ decreased from approximately 0.603 in-domain to 0.148 off-domain, corresponding to a relative reduction of about 75.6\%. This confirms that weed detectors learn dataset-dependent representations and are strongly affected by domain shift. The transfer patterns were also asymmetric, indicating that the generalization ability depends not only on the amount of training data, but also on the diversity, relevance, annotation style, weed-species coverage, and visual similarity between source and target domains.

The multi-source training experiment showed that combining heterogeneous datasets can improve robustness on several target datasets, particularly by exposing the model to a broader range of weed appearances, crop backgrounds, and acquisition conditions. However, the improvements were not uniform across all datasets. ACRE remained challenging, and the multi-source model did not substantially improve over the dataset-specific result on CottonWeedDet12. These findings suggest that multi-source training is useful for improving generality, but it does not fully eliminate domain shift. The relevance and compatibility of the source data remain important for target-domain performance. Overall, the results show that YOLO26 detectors are effective for in-domain weed detection, but robust deployment in unseen field conditions remains challenging. Model scaling alone is not sufficient to overcome domain shift; instead, reliable real-world deployment requires diverse and representative training data, careful dataset selection, target-domain adaptation, and possibly semi-supervised or domain-generalization strategies. Future work will extend the benchmark toward embedded deployment, model compression, domain adaptation, active learning, and the use of unlabelled field imagery to improve generalization across crops, weed species, growth stages, and field conditions.


\bibliographystyle{splncs04}
\bibliography{bibliography}
\end{document}